\def\year{2017}

\documentclass[letterpaper]{article}

\pdfoutput=1 

\usepackage{aaai}
	
\usepackage{times}
\usepackage{helvet}
\usepackage{courier}
\usepackage{latexsym}

\usepackage{amsfonts,eucal,amsbsy,amsopn}
\usepackage{amsmath,amssymb,amsthm,stmaryrd,color}
\usepackage{graphicx}
\usepackage{multirow,rotating}
\usepackage{xspace}
\usepackage{dsfont}
\usepackage{url}
\usepackage{verbatim}
\usepackage[font=small]{caption}
\usepackage{mdframed}

\usepackage{subcaption}
\usepackage{subfloat}
\usepackage{wrapfig}
\usepackage{tikz}

\usepackage{flushend}

\usepackage{tabularx}
\usepackage{booktabs}

\newcommand{\finalversion}[1]{}

\newcommand{\myciteauthoryear}[1]{\citeauthor{#1} (\citeyear{#1})}

\newtheorem{lemma-ap}{Lemma}

\newcommand{\ignore}[1]{}

\def\figref#1{Fig.~\ref{#1}}

\begin{document}

\title{Coherent Dialogue with Attention-based Language Models}

\author{
Hongyuan Mei\\ Johns Hopkins University\\ {\tt hmei@cs.jhu.edu}
\And
Mohit Bansal\\ UNC Chapel Hill\\ {\tt mbansal@cs.unc.edu}
\And
Matthew R. Walter\\ TTI-Chicago\\ {\tt mwalter@ttic.edu}
}

\maketitle

\begin{abstract}
    We model coherent conversation continuation via RNN-based dialogue
    models equipped with a dynamic attention mechanism.  Our
    attention-RNN language model dynamically increases the scope of
    attention on the history as the conversation continues, as opposed
    to standard attention (or alignment) models with a fixed input
    scope in a sequence-to-sequence model. This allows each generated
    word to be associated with the most relevant words in its
    corresponding conversation history.  We evaluate the
    model on two popular dialogue datasets, the open-domain {\it
      MovieTriples} dataset and the closed-domain {\it Ubuntu
      Troubleshoot} dataset, and achieve significant improvements over the state-of-the-art and baselines on several metrics, including complementary diversity-based metrics, human evaluation, and qualitative visualizations.  We also show
    that a vanilla RNN with dynamic attention outperforms more complex
    memory models (e.g., LSTM and GRU) by allowing for flexible, long-distance memory.  We promote further coherence via topic
    modeling-based reranking.
\end{abstract}

\section{Introduction} \label{sec:introduction}

Automatic conversational models~\cite{winograd-71}, also known as dialogue systems, are of great importance to a large variety of applications, ranging from open-domain entertaining chatbots to goal-oriented technical support agents. 
An increasing amount of research has recently been done to build purely data-driven dialogue systems that learn from large corpora of human-to-human conversations, without using hand-crafted rules or templates.
While most work in this area formulates dialogue modeling in a sequence-to-sequence framework (similar to machine translation)~\cite{ritter-11-gen,shang-15-conv,vinyals-15-conv,sordoni-15,li-16-diversity,dusek-16}, some more recent work~\cite{serban-16,luan-16} instead trains a language model over the entire dialogue as one single sequence. In our work, we empirically demonstrate that a language model is better suited to dialogue modeling, as it learns how the conversation evolves as information progresses. Sequence-to-sequence models, on the other hand, learn only how the most recent dialogue response is generated. Such models are better suited to converting the same information from one modality to another, e.g., in machine translation and image captioning.

We improve the coherence of such neural dialogue language models by developing a generative dynamic attention mechanism that allows each generated word to choose which related words it wants to align to in the increasing conversation history (including the previous words in the response being generated). 
Neural attention (or alignment) has proven very successful for various
sequence-to-sequence tasks by associating salient items in the source
sequence with the generated item in the target
sequence~\cite{mnih-14,bahdanau-15,xu-15,mei-16,parikh-16}. However, such
attention models are limited to a fixed scope of
history, corresponding to the input source sequence. In contrast, we
introduce a dynamic attention mechanism to a recurrent neural network
(RNN) language model in which the scope of attention increases  as the recurrence operation progresses from the start through the end of the conversation.
\begin{figure*}[ht] 
  \begin{subfigure}[b]{0.5\linewidth}
    \centering
    \includegraphics[width=0.85\linewidth]{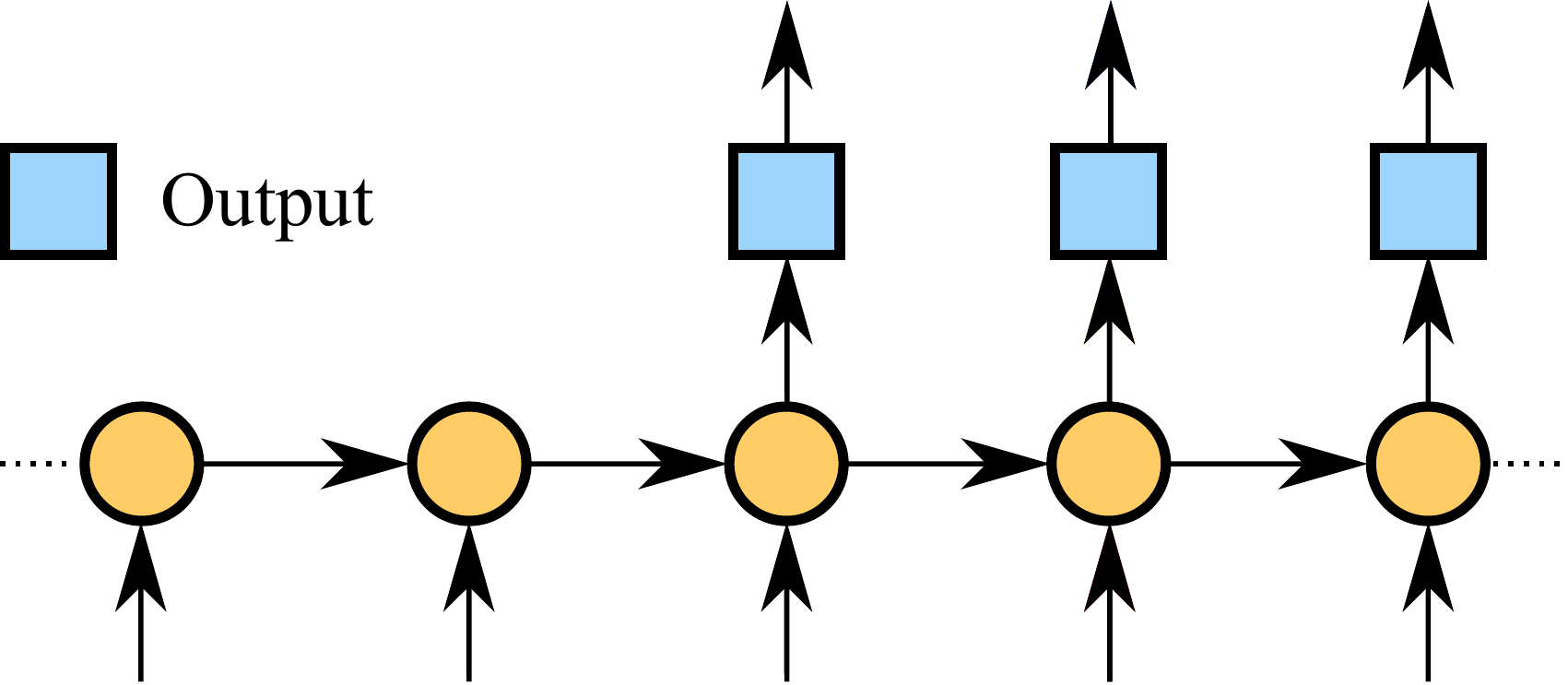} 
    \caption{RNN seq2seq (encoder-decoder) model}\label{fig:seq2seq}%
    \vspace{5pt}
  \end{subfigure}
  \begin{subfigure}[b]{0.5\linewidth}
    \centering
    \includegraphics[width=0.85\linewidth]{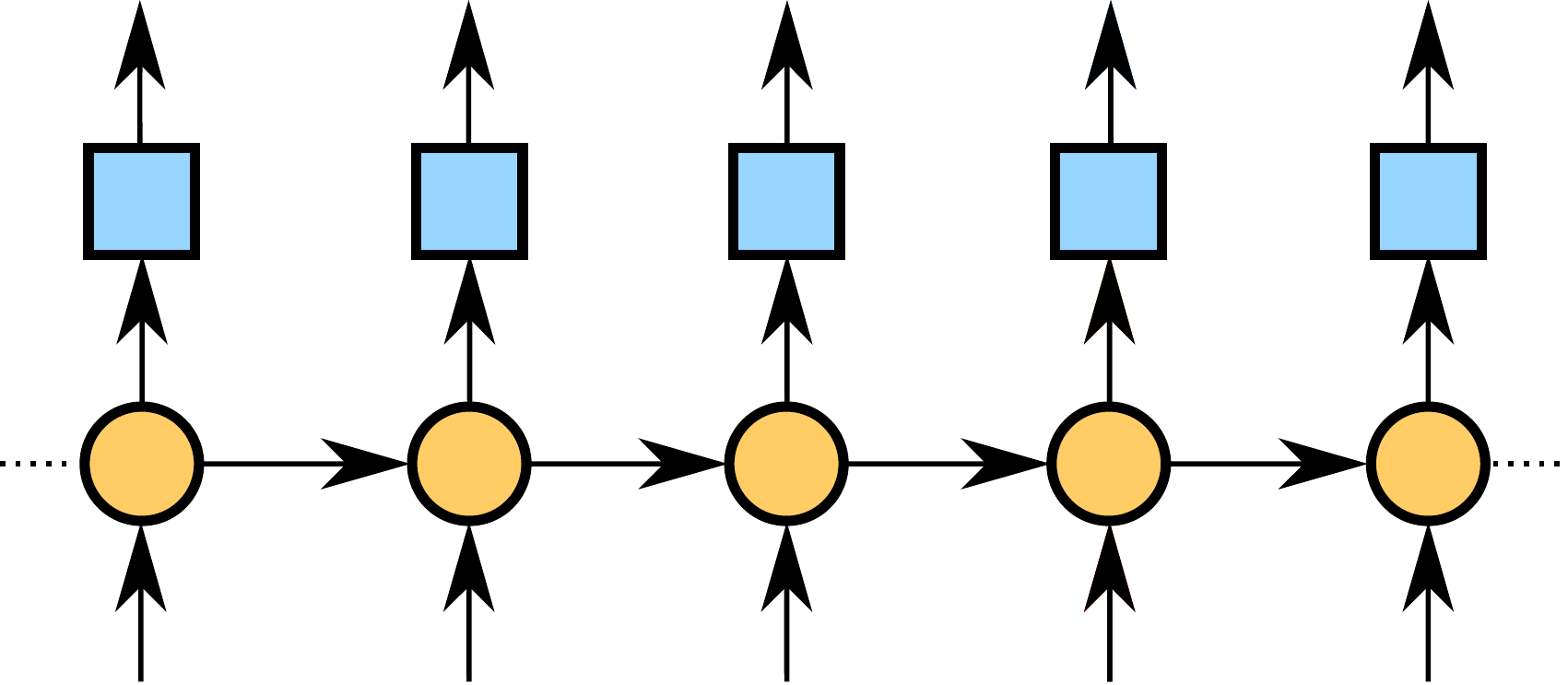}%
    \caption{RNN language model}\label{fig:lang} 
\vspace{5pt}
  \end{subfigure} 
  \begin{subfigure}[b]{0.5\linewidth}
    \centering
    \includegraphics[width=0.85\linewidth]{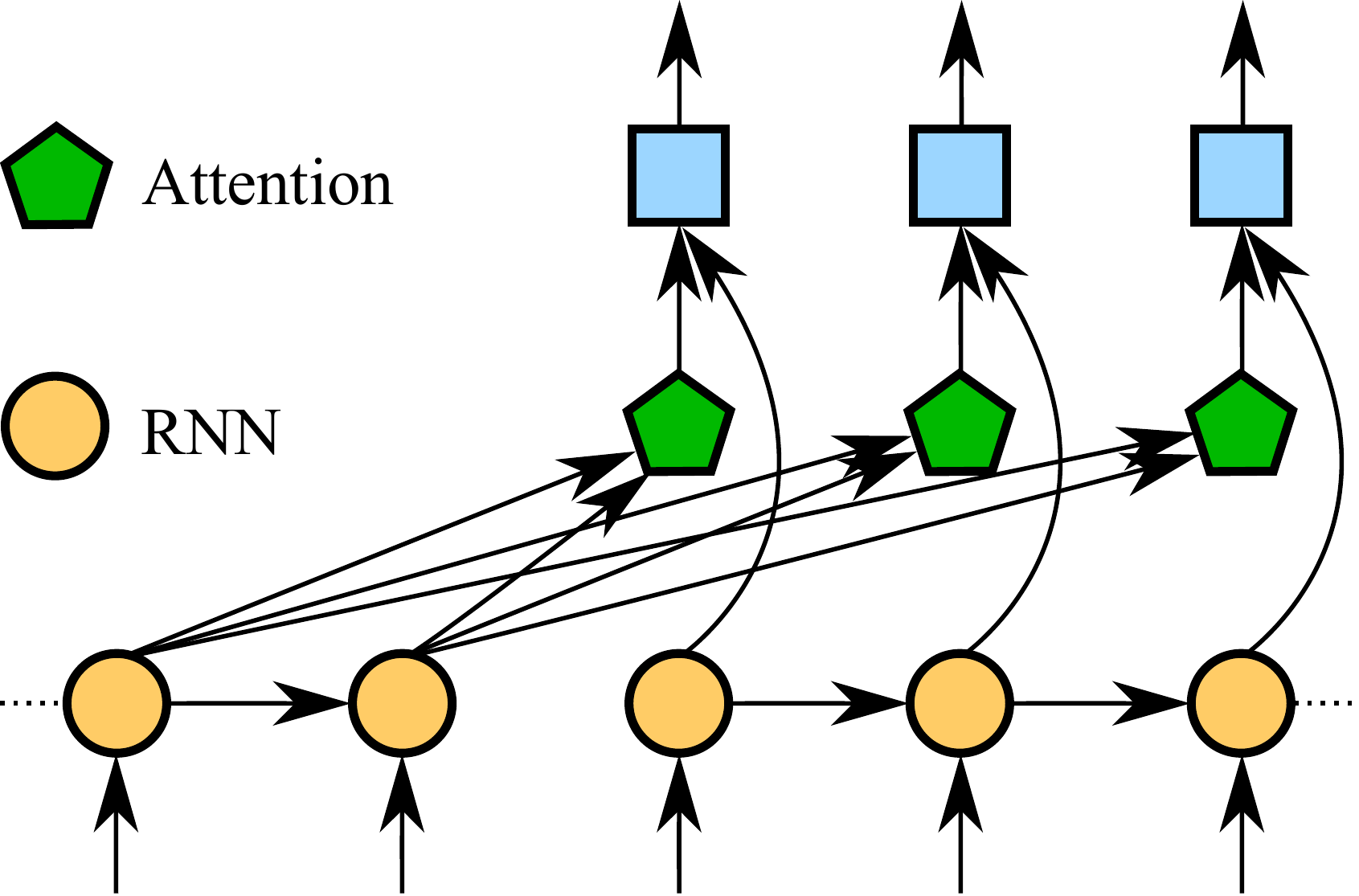} 
    \caption{Attention seq2seq (encoder-decoder) model} 
    \label{fig:att_seq} 
  \end{subfigure}
  \begin{subfigure}[b]{0.5\linewidth}
    \centering
    \includegraphics[width=0.85\linewidth]{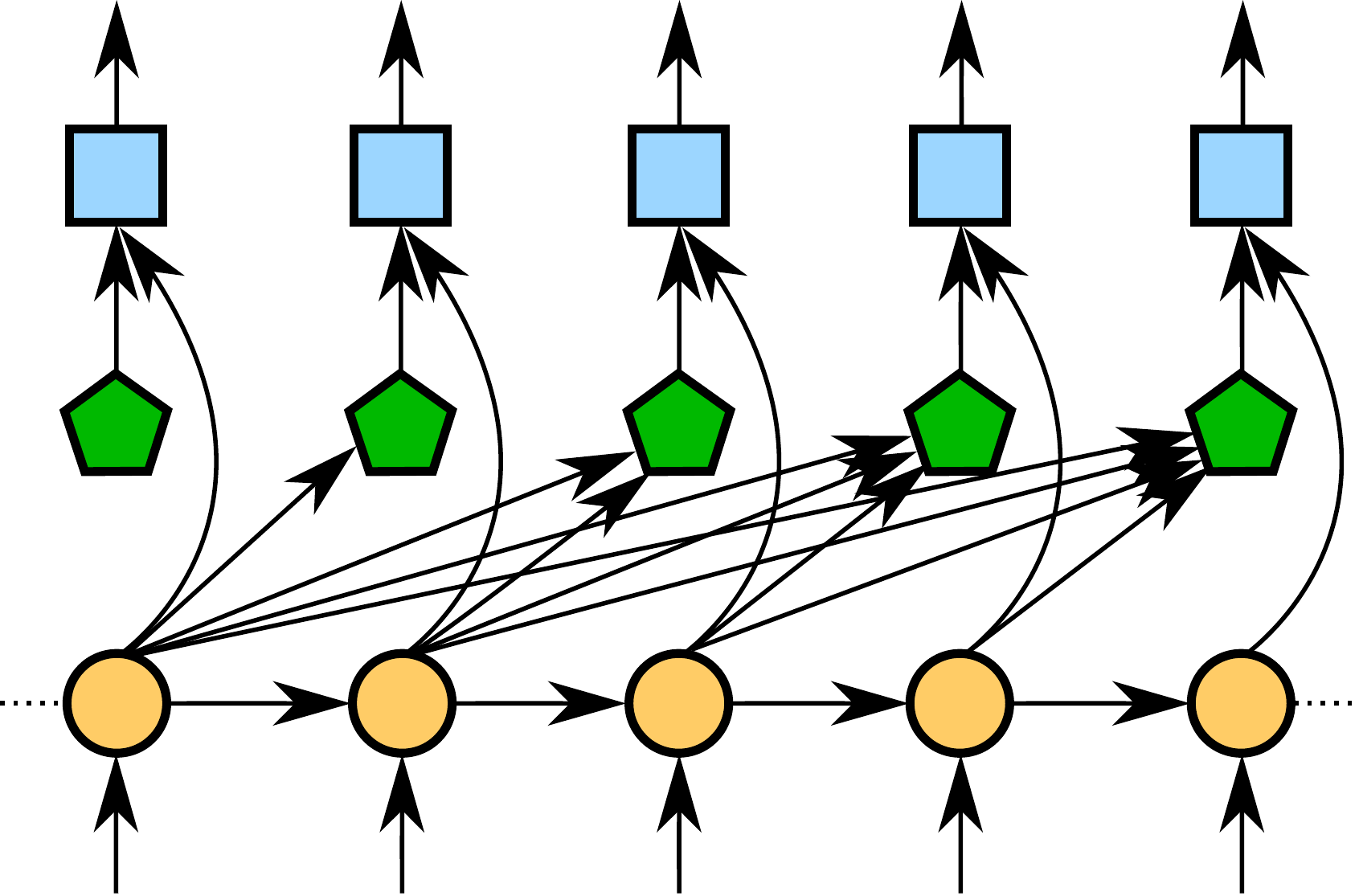} 
    \caption{Attention language model} 
    \label{fig:att_lang} 
  \end{subfigure} 
  \caption{Comparing RNN language models to RNN sequence-to-sequence model, with and without attention.}
  \label{fig7} 
\end{figure*}

The dynamic attention model promotes coherence of the generated dialogue responses (continuations) by favoring the generation of words that have syntactic or semantic associations with salient words in the conversation history. Our simple model shows significant improvements over state-of-the-art models and baselines on several metrics (including complementary diversity-based metrics, human evaluation, and qualitative visualizations) for two benchmark datasets, the open-domain {\it MovieTriples} and closed-domain {\it Ubuntu Troubleshoot} datasets. 
Our vanilla RNN model with dynamic attention outperforms more complex memory models (e.g., LSTM and GRU) by allowing for long-distance and flexible memory. We also present several visualizations to intuitively understand what the attention model is learning. Finally, we also explore a complementary LDA-based method to re-rank the outputs of the soft alignment-based coherence method, further improving performance on the evaluation benchmarks.

\section{Related Work} \label{sec:related}

A great deal of attention has been paid to developing data-driven
methods for natural language dialogue generation.
Conventional statistical approaches tend to rely extensively on
hand-crafted rules and templates,
require interaction with humans or simulated users to optimize parameters,
or produce conversation responses in an information retrieval fashion. 
Such properties prevent training on the large human conversational corpora that are becoming increasingly available, or fail to produce novel natural language responses.

\myciteauthoryear{ritter-11-gen} formulate dialogue response generation as a statistical phrase-based machine translation problem, which requires no explicit hand-crafted rules. 
The recent success of RNNs in statistical machine translation~\cite{sutskever-14,bahdanau-15} has inspired the application of such models to the field of dialogue modeling.
\myciteauthoryear{vinyals-15-conv}
and~\myciteauthoryear{shang-15-conv} employ an RNN to generate responses in  human-to-human conversations by treating the conversation history as one single temporally ordered sequence. 
In such models, the distant relevant context in the history is difficult to recall.
Some efforts have been made to overcome this limitation. \myciteauthoryear{sordoni-15} separately encode the most recent message and all the previous context using a bag-of-words representation, which is decoded using an RNN. 
This approach equates the distance of each word in the generated output to all the words in the conversation history, but loses the temporal information of the history. 
\myciteauthoryear{serban-16} design a hierarchical model that stacks an utterance-level RNN on a token-level RNN, where the utterance-level RNN reduces the number of computational steps between utterances.
\myciteauthoryear{wen-15-semantic} and~\myciteauthoryear{wen-16-multi} improve spoken dialog systems via multi-domain and semantically conditioned neural networks on dialog act representations and explicit slot-value formulations. 

Our work explores the ability of recurrent neural network language models~\cite{bengio-03-lang,mikolov-10-lang} to interpret and generate natural language conversations while still maintaining a relatively simple architecture.
We show that a language model approach outperforms the sequence-to-sequence model at dialogue modeling.
Recently,~\myciteauthoryear{tran-16} demonstrated that the neural
attention mechanism can improve the effectiveness of a neural language
model. We propose an attention-based neural language model for
dialogue modeling that learns how a conversation evolves as a whole,
rather than only how the most recent response is generated, and that also reduces the number of computations between the current recurrence step and the distant relevant context in the conversation history.

The attention mechanism in our model has the additional benefit of favoring words that have semantic association with salient words in the conversation history, which promotes the \emph{coherence} of the topics in the continued dialogue. 
This is important when conversation participants inherently want to maintain the topic of the discussion.
Some past studies have equated coherence with propositional consistency~\cite{goldberg-83}, while others see it as a summary impression~\cite{sanders-83}. Our work falls in the category of viewing coherence as topic continuity~\cite{crow-83,sigman-83}. 
Similar objectives, i.e., generating dialogue responses with certain properties, have been addressed recently, such as promoting response diversity~\cite{li-16-diversity}, enhancing personal consistency~\cite{li-16-persona}, and improving specificity~\cite{yao-16-att}.
Concurrent with this work,~\myciteauthoryear{luan-16} improve topic consistency by feeding into the model the learned LDA-based topic representations. We show that the simple attention neural language model significantly outperforms such a design.
Furthermore, we suggest an LDA-based re-ranker complementary to soft neural attention that further promotes topic coherence.


\section{The Model}
\label{sec:model}

In this section, we introduce an attention-RNN dialogue model and compare it to the architectures of basic RNN and sequence-to-sequence models.

\subsection{RNN Seq2Seq and  Language Models}
\label{sec:rnn-lm}

Recurrent neural networks have been successfully used 
both in sequence-to-sequence models
(RNN-Seq2Seq,~\figref{fig:seq2seq})~\cite{sutskever-14} and in language
models
(RNN-LM,~\figref{fig:lang})~\cite{bengio-03-lang,mikolov-10-lang}. We first
discuss language models for dialogue, which is the primary focus of our
work, then briefly introduce the sequence-to-sequence model, and lastly
discuss the use of attention methods in both models.

The RNN-LM models a sentence as a sequence of tokens
$\{ w_0, w_1, \ldots, w_T \}$ with a recurrence function
\begin{equation}
    h_t = f(h_{t-1}, w_{t-1}) \label{equ:ht}
\end{equation}
and an output (softmax) function
\begin{equation}
  P(w_{t} = v_j \vert w_{0:t-1}) = \frac{\exp g(h_{t}, v_j)}{\sum_{i} \exp g(h_{t}, v_i)},
\end{equation}
where the recurrent hidden state $h_{t} \in \mathbb{R}^d$ encodes all the tokens up to $t-1$ and is used to compute the probability of generating $v_j \in V$ as the next token from the vocabulary $V$.

The functions $f$ and $g$ are typically defined as
\begin{subequations}
    \begin{align}
      f(h_{t-1}, w_{t-1}) &= \tanh ( H h_{t-1} + P E_{w_{t-1}} ) \label{equ:fdef} \\
      g(h_{t}, v_{j}) &= O_{v_j}^\top h_{t},
    \end{align}
\end{subequations}
where $H \in \mathbb{R}^{d \times d}$ is the recurrence matrix, 
$E_{w_{t-1}}$ is a column of word embedding matrix \mbox{$E \in \mathbb{R}^{d_e \times V}$} that corresponds to $w_{t-1}$, 
$P \in \mathbb{R}^{d \times d_e}$ projects word embedding into the space of the same dimension $d$ as the hidden units, 
and $O \in \mathbb{R}^{d \times V}$ is the output word embedding matrix
with column vector $O_{v_j}$ corresponding to $v_j$.  

We train the RNN-LM, i.e, estimate the parameters $H$, $P$, $E$ and $O$, by maximizing the log-likelihood on a set of natural language training sentences of size $N$
\begin{equation}
    \ell = \frac{1}{N} \sum_{n=1}^N \sum_{t=0}^{T_n} \log P(w_t \vert w_{0:t-1})
\end{equation}
Since the entire architecture is differentiable, the objective can be optimized by back-propagation. 

When dialogue is formulated as a sequence-to-sequence task,
the RNN-Seq2Seq model can be used in order to predict
a target sequence \mbox{$w_{0:L}^T = \{ w_0^T, w_1^T, \ldots, w_{L}^T \}$}
given an input source sequence 
\mbox{$w_{0:M}^S = \{ w_0^S, w_1^S, \ldots, w_{M}^S \}$}.
In such settings, an encoder RNN represents the input as a sequence of hidden states
\mbox{$h_{0:M}^S = \{h_0^S, h_1^S, \ldots, h_M^S\}$}, 
and a separate decoder RNN then predicts the target sequence token-by-token given the encoder hidden states $h_{0:M}^S$.

\subsection{Attention in RNN-Seq2Seq Models}

There are several ways by which to integrate the sequence of hidden states $h_{0:M}^S$ in the decoder RNN. An attention mechanism (Fig.~\ref{fig:att_seq}) has proven to be particularly effective for various related tasks in machine translation, image caption synthesis, and language understanding~\cite{mnih-14,bahdanau-15,xu-15,mei-16}.

The attention module takes as input the encoder hidden state sequence $h_{0:M}^S$ and the decoder hidden state $h_{l-1}^T$ at each step $l-1$, 
and returns a context vector $z_{l}$ computed as a weighted average of
encoder hidden states $h_{0:M}^S$
\begin{subequations}
    \begin{align}
      \beta_{lm} &= b^\top \tanh (W h_{l-1}^T + U h_{m}^S) \\
      \alpha_{lm} &= \exp (\beta_{lm}) / \sum_{m=0}^M \exp (\beta_{lm}) \\
      z_{l} &= \sum_{m=0}^M \alpha_{lm} h_{m}^S,
    \end{align}
\end{subequations}
where parameters $W \in \mathbb{R}^{d \times d}$, $U \in \mathbb{R}^{d \times d}$, and \mbox{$b \in \mathbb{R}^{d}$} are jointly learned with the other model parameters. 
The context vector $z_{l}$ is then used as an extra input to the decoder RNN at step $l$ together with $w_{0:l-1}^T$ to predict the next token $w_{l}^T$. 

\subsection{Attention in RNN-LM}
\label{sec:a-rnn-lm}

We develop an attention-RNN language model (A-RNN-LM) as illustrated in
Figure~\ref{fig:att_lang}, and describe how it can be used in the context
of dialogue modeling. We then describe its advantages compared to the use
of attention in sequence-to-sequence models. 

As with the RNN-LM, the model first encodes the input into a sequence of hidden states up to word $t-1$ (Eqn.~\ref{equ:ht}). Given a representation of tokens up to $t-1$ $\{r_0, r_1, \ldots, r_{t-1}\}$ (which we define shortly), 
the attention module computes the context vector $z_t$ at step $t$ as a weighted average of ${r_{0:{t-1}}}$ 
\begin{subequations}
    \begin{align}
      \beta_{ti} &= b^\top \tanh (W h_{t-1} + U r_{i}) \\
      \alpha_{ti} &= \exp (\beta_{ti}) / \sum_{i=0}^{t-1} \exp (\beta_{ti})\\
      z_{t} &= \sum_{i=0}^{t-1} \alpha_{ti} r_{i}
    \end{align}
\end{subequations}
We then use the context vector $z_t$ together with the hidden state $h_t$ to predict the output at time $t$
\begin{subequations} 
    \begin{align}
      g(h_t, z_t, v_j) &= O_{v_j}^\top (O_h h_t + O_z z_t ) \\
      P(w_{t} = v_j | w_{0:t-1}) &= \frac{\exp g(h_{t}, z_t, v_j)}{\sum_{i} \exp g(h_{t}, z_t, v_i)},
    \end{align}
\end{subequations}
where $O_h  \in \mathbb{R}^{d \times d}$ and $O_z \in \mathbb{R}^{d \times d_z}$ project $h_t$ and $z_t$, respectively, into the same space of dimension $d$. 

There are multiple benefits of using an attention-RNN language model for
dialogue, which are empirically supported by our experimental
results. First, a complete dialogue is usually composed of multiple turns. A language model over the entire dialogue is expected to
better learn how a conversation evolves as a whole, unlike a
sequence-to-sequence model, which only learns how the most recent response
is generated and is better suited to translation-style tasks that
transform the same information from one modality to another. Second,
compared to LSTM models, an attention-based RNN-LM also allows for gapped
context and a flexible combination of conversation history for
every individual generated token, while maintaining low model complexity. Third, attention models yield  interpretable results---we visualize the learned attention weights, showing how attention chooses the salient words from the dialogue history that are important for generating each new word. 
Such a visualization is typically harder for the hidden states and gates of conventional LSTM and RNN language models. 

With an attention mechanism, there are multiple options for defining the token representation $r_{0:t-1}$. The original attention model introduced by \myciteauthoryear{bahdanau-15} uses the hidden units $h_{0:t-1}$ as the token representations $r_{0:t-1}$. 
Recent work~\cite{mei-16,mei-16-gen} has demonstrated that performance can be improved by using multiple abstractions of the input, e.g., \mbox{$r_{i} = (E_{w_{i}}^\top, h_{i}^\top)^\top$}, which is what we use in this work.

\subsection{LDA-based Re-Ranking}
\label{sec:lda-rerank}

While the trained attention-RNN dialogue model generates a natural language continuation of a conversation while maintaining topic concentration by token association, some dialogue-level topic-supervision can help to encourage generations that are more topic-aware.
Such supervision is not commonly available, and we use unsupervised methods
to learn document-level latent topics. We employ the learned topic model to select the best continuation based on document-level topic-matching. 

We choose Latent Dirichlet Allocation (LDA)~\cite{blei-03-lda,blei-09-topic}
due to its demonstrated ability to learn a distribution over latent topics given a collection of documents. 
This generative model assumes documents $\{w_{0:T_n}\}_{n=1}^{N}$ arise from $K$ topics, each of which is defined as a distribution over a fixed vocabulary of terms, which forms a graphical structure $\mathcal{L}$ that can be learned from the training data.
The topic representation $\hat{\theta}$ of a (possibly unseen) dialogue $w_{0:T}$ can then be estimated with the learned topic structure $\mathcal{L}$ as $\hat{\theta}(w_{0:T}) = \mathcal{L}(w_{0:T})$.

Given a set of generated continuations $\{c_m\}_{m=1}^M$ for each unseen dialogue $w_{0:T}$, the topic representations of the dialogue and its continuations are $\hat{\theta}(w_{0:T}) = \mathcal{L}(w_{0:T})$ and $\hat{\theta}(c_m) = \mathcal{L}(c_m)$, respectively.
We employ a matching score $\mathcal{S}_m =
\mathcal{S}\bigl(\hat{\theta}(w_{0:T}), \hat{\theta}(c_m)\bigr)$ to compute
the similarity between  $\hat{\theta}(w_{0:T})$ and each $\hat{\theta}(c_m)$. 
In the end, a weighted score is computed as 
\mbox{$\bar{\mathcal{S}}_m = \lambda \mathcal{S}_m + (1-\lambda) \ell(c_m|w_{0:T})$}, 
where $\lambda \in [0, 1]$
and $\ell(c_m|w_{0:T})$ is the conditional log-likelihood of the continuation $c_m$.
The hyper-parameters $K$ and $\lambda$ are tuned on a development set.

Concurrent with our work,~\myciteauthoryear{luan-16} use learned topic representations $\hat{\theta}$ of the given conversation as an extra feature in a language model to enhance the topic coherence of the generation.
As we show in the Results and Analysis section,
our model significantly outperforms this approach.

\section{Experimental Setup}
\label{sec:experimentalsetup}

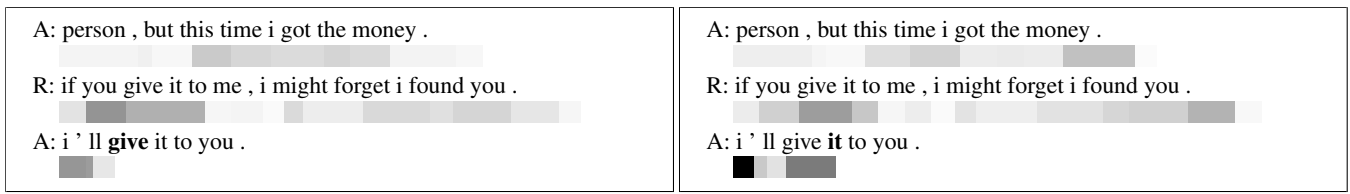
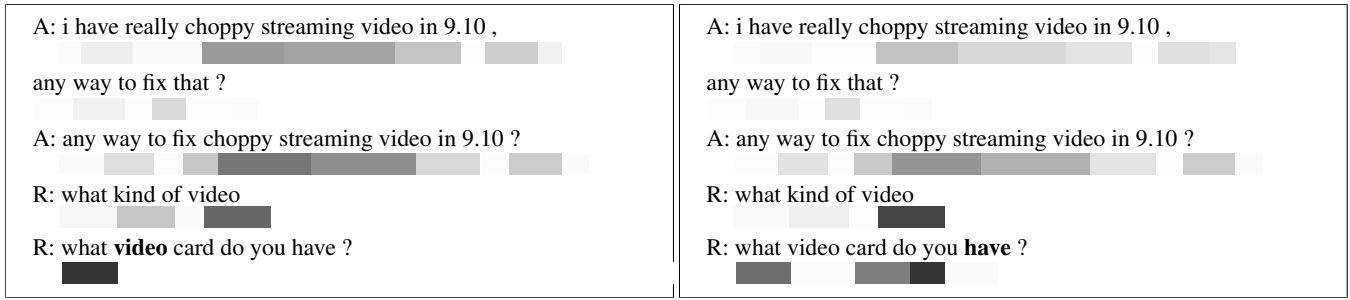
\begin{figure*}[!tb]
	\begin{subfigure}[b]{1.0\linewidth}
	\begin{minipage}[t]{1.0\linewidth}
	{\small%
      \begin{minipage}[t]{0.5\linewidth}
     	\begin{mdframed}[align=center]{\small
        	A: person , but this time i got the money .\\
        	\begin{tikzpicture}[xscale=1.0,yscale=0.8]
        		\draw[step=1pt,white,very thin] (0,0) grid (170pt,10pt);
        		\fill[black!4.5!white] (10pt,0) rectangle (40pt,10pt);
        		\fill[black!6.0!white] (40pt,0) rectangle (45pt,10pt);
        		\fill[black!3.0!white] (45pt,0) rectangle (60pt,10pt);
        		\fill[black!20.6!white] (60pt,0) rectangle (75pt,10pt);
        		\fill[black!16.0!white] (75pt,0) rectangle (90pt,10pt);
        		\fill[black!14.0!white] (90pt,0) rectangle (110pt,10pt);
        		\fill[black!17.0!white] (110pt,0) rectangle (120pt,10pt);
        		\fill[black!14.5!white] (120pt,0) rectangle (135pt,10pt);
        		\fill[black!4.8!white] (135pt,0) rectangle (160pt,10pt);
        		\fill[black!3.0!white] (160pt,0) rectangle (170pt,10pt);
        	\end{tikzpicture} \\
           R: if you give it to me , i might forget i found you .\\
           \begin{tikzpicture}[xscale=1.0,yscale=0.8]
        		\draw[step=1pt,white,very thin] (0,0) grid (207pt,10pt);
        		\fill[black!11.0!white] (10pt,0) rectangle (20pt,10pt);
        		\fill[black!42.0!white] (20pt,0) rectangle (35pt,10pt);
        		\fill[black!31.0!white] (35pt,0) rectangle (55pt,10pt);
        		\fill[black!31.0!white] (55pt,0) rectangle (65pt,10pt);
        		\fill[black!3.5!white] (65pt,0) rectangle (75pt,10pt);
        		\fill[black!4.4!white] (75pt,0) rectangle (87pt,10pt);
        		\fill[black!1.8!white] (87pt,0) rectangle (95pt,10pt);
        		\fill[black!14.5!white] (95pt,0) rectangle (102pt,10pt);
        		\fill[black!7.0!white] (102pt,0) rectangle (125pt,10pt);
        		\fill[black!15.0!white] (125pt,0) rectangle (150pt,10pt);
        		\fill[black!12.0!white] (150pt,0) rectangle (159pt,10pt);
        		\fill[black!16.5!white] (159pt,0) rectangle (181pt,10pt);
        		\fill[black!10.0!white] (181pt,0) rectangle (199pt,10pt);
        		\fill[black!3.5!white] (199pt,0) rectangle (207pt,10pt);
        	\end{tikzpicture} \\
           A: i ' ll \textbf{give} it to you .\\
           \begin{tikzpicture}[xscale=1.0,yscale=0.8]
        		\draw[step=1pt,white,very thin] (0,0) grid (100pt,10pt);
        		\fill[black!41.0!white] (10pt,0) rectangle (20pt,10pt);
        		\fill[black!39.0!white] (20pt,0) rectangle (23pt,10pt);
        		\fill[black!9.5!white] (23pt,0) rectangle (31pt,10pt);
        		\fill[black!0!white] (31pt,0) rectangle (50pt,10pt);
        		\fill[black!0!white] (50pt,0) rectangle (60pt,10pt);
        		\fill[black!0!white] (60pt,0) rectangle (70pt,10pt);
        		\fill[black!0!white] (70pt,0) rectangle (85pt,10pt);
        		\fill[black!0!white] (85pt,0) rectangle (90pt,10pt);
        	\end{tikzpicture}}
       	\end{mdframed}
	\end{minipage}}
  {\small
	\begin{minipage}[t]{0.5\linewidth}
     	\begin{mdframed}[align=center]{\small
        	A: person , but this time i got the money .\\
        	\begin{tikzpicture}[xscale=1.0,yscale=0.8]
        		\draw[step=1pt,white,very thin] (0,0) grid (170pt,10pt);
        		\fill[black!6.8!white] (10pt,0) rectangle (40pt,10pt);
        		\fill[black!4.1!white] (40pt,0) rectangle (45pt,10pt);
        		\fill[black!2.6!white] (45pt,0) rectangle (60pt,10pt);
        		\fill[black!12.8!white] (60pt,0) rectangle (77pt,10pt);
        		\fill[black!17.8!white] (77pt,0) rectangle (96pt,10pt);
        		\fill[black!7.3!white] (96pt,0) rectangle (110pt,10pt);
        		\fill[black!8.2!white] (110pt,0) rectangle (120pt,10pt);
        		\fill[black!7.6!white] (120pt,0) rectangle (135pt,10pt);
        		\fill[black!24.4!white] (135pt,0) rectangle (162pt,10pt);
        		\fill[black!1.5!white] (162pt,0) rectangle (170pt,10pt);
        	\end{tikzpicture} \\
           R: if you give it to me , i might forget i found you .\\
           \begin{tikzpicture}[xscale=1.0,yscale=0.8]
        		\draw[step=1pt,white,very thin] (0,0) grid (205pt,10pt);
        		\fill[black!7.4!white] (10pt,0) rectangle (20pt,10pt);
        		\fill[black!18.6!white] (20pt,0) rectangle (35pt,10pt);
        		\fill[black!38.5!white] (35pt,0) rectangle (55pt,10pt);
        		\fill[black!21.8!white] (55pt,0) rectangle (65pt,10pt);
        		\fill[black!3.7!white] (65pt,0) rectangle (75pt,10pt);
        		\fill[black!7.2!white] (75pt,0) rectangle (85pt,10pt);
        		\fill[black!1.9!white] (85pt,0) rectangle (94pt,10pt);
        		\fill[black!11.1!white] (94pt,0) rectangle (102pt,10pt);
        		\fill[black!6.5!white] (102pt,0) rectangle (125pt,10pt);
        		\fill[black!11.4!white] (125pt,0) rectangle (150pt,10pt);
        		\fill[black!16.0!white] (150pt,0) rectangle (160pt,10pt);
        		\fill[black!18.5!white] (160pt,0) rectangle (182pt,10pt);
        		\fill[black!30.0!white] (182pt,0) rectangle (200pt,10pt);
        		\fill[black!2.9!white] (200pt,0) rectangle (210pt,10pt);
        	\end{tikzpicture} \\
           A: i ' ll give \textbf{it} to you .\\
           \begin{tikzpicture}[xscale=1.0,yscale=0.8]
        		\draw[step=1pt,white,very thin] (0,0) grid (100pt,10pt);
        		\fill[black!100.0!white] (10pt,0) rectangle (18pt,10pt);
        		\fill[black!21.0!white] (18pt,0) rectangle (23pt,10pt);
        		\fill[black!11.5!white] (23pt,0) rectangle (30pt,10pt);
        		\fill[black!51.9!white] (30pt,0) rectangle (49pt,10pt);
        		\fill[black!0!white] (49pt,0) rectangle (60pt,10pt);
        		\fill[black!0!white] (60pt,0) rectangle (70pt,10pt);
        		\fill[black!0!white] (70pt,0) rectangle (85pt,10pt);
        		\fill[black!0!white] (85pt,0) rectangle (90pt,10pt);
        	\end{tikzpicture}}
       	\end{mdframed}
	\end{minipage}}
    \end{minipage}
    \caption{{\it MovieTriples}} 
    \label{att:movie} 
    \vspace{5pt}
    \end{subfigure}
    \begin{subfigure}[b]{1.0\linewidth}
	\begin{minipage}[t]{1.0\linewidth}
        {\small
	\begin{minipage}[t]{0.5\linewidth}
     	\begin{mdframed}[align=center]{\small
        	A: i have really choppy streaming video in 9.10 , \\
        	\begin{tikzpicture}[xscale=1.0,yscale=0.8]
        		\draw[step=1pt,white,very thin] (0,0) grid (210pt,10pt);
        		\fill[black!1.4!white] (10pt,0) rectangle (18pt,10pt);
        		\fill[black!6.5!white] (18pt,0) rectangle (38pt,10pt);
        		\fill[black!2.12!white] (38pt,0) rectangle (64pt,10pt);
        		\fill[black!39.8!white] (64pt,0) rectangle (95pt,10pt);
        		\fill[black!36.0!white] (95pt,0) rectangle (137pt,10pt);
        		\fill[black!22.9!white] (137pt,0) rectangle (162pt,10pt);
        		\fill[black!0.4!white] (162pt,0) rectangle (171pt,10pt);
        		\fill[black!19.2!white] (171pt,0) rectangle (191pt,10pt);
        		\fill[black!5.4!white] (191pt,0) rectangle (200pt,10pt);
        	\end{tikzpicture} \\
        	any way to fix that ?\\
        	\begin{tikzpicture}[xscale=1.0,yscale=0.8]
        		\draw[step=1pt,white,very thin] (0,0) grid (210pt,10pt);
        		\fill[black!0.9!white] (0pt,0) rectangle (15pt,10pt);
        		\fill[black!5.62!white] (15pt,0) rectangle (35pt,10pt);
        		\fill[black!1.0!white] (35pt,0) rectangle (45pt,10pt);
        		\fill[black!14.8!white] (45pt,0) rectangle (58pt,10pt);
        		\fill[black!1.0!white] (58pt,0) rectangle (75pt,10pt);
        		\fill[black!1.6!white] (75pt,0) rectangle (85pt,10pt);
        	\end{tikzpicture} \\
           A: any way to fix choppy streaming video in 9.10 ?\\
           \begin{tikzpicture}[xscale=1.0,yscale=0.8]
        		\draw[step=1pt,white,very thin] (0,0) grid (210pt,10pt);
        		\fill[black!2.1!white] (10pt,0) rectangle (27pt,10pt);
        		\fill[black!13.0!white] (27pt,0) rectangle (46pt,10pt);
        		\fill[black!1.2!white] (46pt,0) rectangle (57pt,10pt);
        		\fill[black!21.4!white] (57pt,0) rectangle (70pt,10pt);
        		\fill[black!53.7!white] (70pt,0) rectangle (105pt,10pt);
        		\fill[black!44.4!white] (105pt,0) rectangle (145pt,10pt);
        		\fill[black!15.3!white] (145pt,0) rectangle (169pt,10pt);
        		\fill[black!0.9!white] (170pt,0) rectangle (180pt,10pt);
        		\fill[black!19.9!white] (180pt,0) rectangle (200pt,10pt);
        		\fill[black!1.9!white] (200pt,0) rectangle (210pt,10pt);
        	\end{tikzpicture} \\
           R: what kind of video\\
           \begin{tikzpicture}[xscale=1.0,yscale=0.8]
        		\draw[step=1pt,white,very thin] (0,0) grid (210pt,10pt);
        		\fill[black!3.7!white] (10pt,0) rectangle (32pt,10pt);
        		\fill[black!22.5!white] (32pt,0) rectangle (54pt,10pt);
        		\fill[black!2.0!white] (54pt,0) rectangle (65pt,10pt);
        		\fill[black!59.8!white] (65pt,0) rectangle (90pt,10pt);
        	\end{tikzpicture} \\
           R: what \textbf{video} card do you have ?\\
           \begin{tikzpicture}[xscale=1.0,yscale=0.8]
        		\draw[step=1pt,white,very thin] (0,0) grid (210pt,10pt);
        		\fill[black!79.4!white] (11pt,0) rectangle (32pt,10pt);
        		\fill[black!0!white] (32pt,0) rectangle (55pt,10pt);
        		\fill[black!0!white] (55pt,0) rectangle (750pt,10pt);
        		\fill[black!0!white] (75pt,0) rectangle (90pt,10pt);
        		\fill[black!0!white] (90pt,0) rectangle (110pt,10pt);
        		\fill[black!0!white] (110pt,0) rectangle (130pt,10pt);
        		\fill[black!0!white] (130pt,0) rectangle (135pt,10pt);
        	\end{tikzpicture}}
       	\end{mdframed}
	\end{minipage}}
  {\small
	\begin{minipage}[t]{0.5\linewidth}
     	\begin{mdframed}[align=center]{\small
        	A: i have really choppy streaming video in 9.10 , \\
        	\begin{tikzpicture}[xscale=1.0,yscale=0.8]
        		\draw[step=1pt,white,very thin] (0,0) grid (210pt,10pt);
        		\fill[black!1.5!white] (10pt,0) rectangle (20pt,10pt);
        		\fill[black!2.6!white] (20pt,0) rectangle (40pt,10pt);
        		\fill[black!0.6!white] (40pt,0) rectangle (64pt,10pt);
        		\fill[black!23.6!white] (64pt,0) rectangle (95pt,10pt);
        		\fill[black!15.5!white] (95pt,0) rectangle (136pt,10pt);
        		\fill[black!10.8!white] (136pt,0) rectangle (161pt,10pt);
        		\fill[black!0.3!white] (161pt,0) rectangle (171pt,10pt);
        		\fill[black!12.3!white] (171pt,0) rectangle (190pt,10pt);
        		\fill[black!10.4!white] (190pt,0) rectangle (200pt,10pt);
        	\end{tikzpicture} \\
        	any way to fix that ?\\
        	\begin{tikzpicture}[xscale=1.0,yscale=0.8]
        		\draw[step=1pt,white,very thin] (0,0) grid (210pt,10pt);
        		\fill[black!0.6!white] (0pt,0) rectangle (15pt,10pt);
        		\fill[black!3.3!white] (15pt,0) rectangle (35pt,10pt);
        		\fill[black!0.4!white] (35pt,0) rectangle (45pt,10pt);
        		\fill[black!12.7!white] (45pt,0) rectangle (58pt,10pt);
        		\fill[black!1.1!white] (58pt,0) rectangle (75pt,10pt);
        		\fill[black!1.7!white] (75pt,0) rectangle (85pt,10pt);
        	\end{tikzpicture} \\
           A: any way to fix choppy streaming video in 9.10 ?\\
           \begin{tikzpicture}[xscale=1.0,yscale=0.8]
        		\draw[step=1pt,white,very thin] (0,0) grid (210pt,10pt);
        		\fill[black!0.9!white] (10pt,0) rectangle (27pt,10pt);
        		\fill[black!11.2!white] (27pt,0) rectangle (46pt,10pt);
        		\fill[black!0.6!white] (46pt,0) rectangle (56pt,10pt);
        		\fill[black!20.8!white] (56pt,0) rectangle (70pt,10pt);
        		\fill[black!42.1!white] (70pt,0) rectangle (104pt,10pt);
        		\fill[black!30.4!white] (104pt,0) rectangle (145pt,10pt);
        		\fill[black!11.1!white] (145pt,0) rectangle (170pt,10pt);
        		\fill[black!0.4!white] (170pt,0) rectangle (180pt,10pt);
        		\fill[black!19.9!white] (180pt,0) rectangle (200pt,10pt);
        		\fill[black!1.8!white] (200pt,0) rectangle (210pt,10pt);
        	\end{tikzpicture} \\
           R: what kind of video\\
           \begin{tikzpicture}[xscale=1.0,yscale=0.8]
        		\draw[step=1pt,white,very thin] (0,0) grid (210pt,10pt);
        		\fill[black!1.9!white] (10pt,0) rectangle (31pt,10pt);
        		\fill[black!6.1!white] (31pt,0) rectangle (54pt,10pt);
        		\fill[black!1.2!white] (54pt,0) rectangle (65pt,10pt);
        		\fill[black!73.1!white] (65pt,0) rectangle (90pt,10pt);
        	\end{tikzpicture} \\
           R: what video card do you \textbf{have} ?\\
           \begin{tikzpicture}[xscale=1.0,yscale=0.8]
        		\draw[step=1pt,white,very thin] (0,0) grid (210pt,10pt);
        		\fill[black!57.0!white] (11pt,0) rectangle (32pt,10pt);
        		\fill[black!1.6!white] (32pt,0) rectangle (56pt,10pt);
        		\fill[black!50.5!white] (56pt,0) rectangle (77pt,10pt);
        		\fill[black!79.8!white] (77pt,0) rectangle (90pt,10pt);
        		\fill[black!1.9!white] (90pt,0) rectangle (110pt,10pt);
        		\fill[black!0!white] (110pt,0) rectangle (130pt,10pt);
        		\fill[black!0!white] (130pt,0) rectangle (135pt,10pt);
        	\end{tikzpicture}}
       	\end{mdframed}
	\end{minipage}}
    \end{minipage}
    \caption{{\it Ubuntu Troubleshoot}} 
    \label{att:ubuntu} 
    \vspace{5pt}
    \end{subfigure}
    \caption{
    	A visualization of attention on the \subref{att:movie} {\it MovieTriples} and \subref{att:ubuntu} {\it Ubuntu Troubleshooting} datasets, showing which words in the conversation history are being aligned to, for each generated response word. 
    	Shaded intervals indicate the strength with which the corresponding words in the conversation history and response are attend to when generating the bolded word in the response.
      We show this for two generated words in the same response (left and right column).
      }
\label{fig:att-viz}
\end{figure*}
\subsection{Dataset}
We train and evaluate the models on two large natural language
dialogue datasets, {\it MovieTriples} (pre-processed
by~\myciteauthoryear{serban-16}) and {\it Ubuntu Troubleshoot}
(pre-processed by~\myciteauthoryear{luan-16}).
The dialogue within each of these datasets consists of a sequence of
utterances (turns), each of which is a sequence of tokens
(words).\footnote{Following 
  \myciteauthoryear{luan-16}, we randomly sample nine utterances as
  negative examples of the last utterance for each conversation in
  {\it Ubuntu Troubleshoot} for the development set.} The accompanying
supplementary material provides the statistics for 
these two datasets.

\subsection{Evaluation Metrics}

For the sake of comparison, we closely follow previous work and adopt
several standard (and complementary) evaluation metrics: perplexity (PPL), word error
rate (WER), recall@N, BLEU, and diversity-based Distinct-1. We provide further discussion of the
various metrics and their advantages in the supplementary material.
On the {\it MovieTriples} dataset, we use PPL and WER, as is done in previous work.
Following \myciteauthoryear{serban-16}, we adopt two versions for each metric:
i) PPL as the word-level perplexity over the entire dialogue conversation; 
ii) PPL@L as the word-level perplexity over the last utterance of the conversation; 
iii) WER; 
and 
iv) WER@L (defined similarly). 

On the {\it Ubuntu} dataset, we follow previous work and use PPL and recall@N.
Recall@N~\cite{manning-08-introduction} evaluates a model by
measuring how often the model ranks the correct dialogue continuation within top-N given $10$ candidates.
Additionally, we also employ the BLEU score~\cite{papineni-01} to evaluate
the quality of the generations produced by the models. 
Following~\myciteauthoryear{luan-16}, 
we perform model selection using PPL on the development set, 
and perform the evaluation on the test set using the other metrics. 
We also present evaluation using the Distinct-1 metric (proposed by~\myciteauthoryear{li-16-diversity}) to measure the ability of the A-RNN to promote diversity in the generations, because typical neural dialogue models generate generic, safe responses (technically appropriate but not informative, e.g., ``I donÕt know'').
Finally, we also present a preliminary human evaluation.

\subsection{Training Details}

For the {\it MovieTriples} dataset, we follow the same procedure as~\myciteauthoryear{serban-16} and first pretrain on the large {\it Q-A SubTitle} dataset
~\cite{ameixa2014luke}, which contains $5.5$M question-answer pairs from which we randomly sample $20000$ pairs as the held-out set, and then fine-tune on the target {\it MovieTriples} dataset. We perform early-stopping according to the PPL score on the held-out set. We train the models for both the {\it MovieTriples} and {\it Ubuntu Troubleshoot} datasets using Adam~\cite{kingma-15} for optimization in RNN back-propagation. The accompanying supplementary material provides additional training details, including the hyper-parameter settings.

\section{Results and Analysis}
\label{sec:results}

\subsection{Primary Dialogue Modeling Results}
\label{sec:dialog-model}

In this section, we compare the performance on several metrics of our attention-based
RNN-LM with RNN baselines and state-of-the-art
models on the two benchmark datasets. Table~\ref{tab:movie-test}
reports PPL and WER results on the {\it MovieTriples} test set, while Table~\ref{tab:ubuntu-ppl-rank} compares different models on {\it Ubuntu Troubleshoot} in terms of PPL on the development set and recall@N ($N=1\ \text{and}\ 2$) on the test set (following what previous work reports).
In the tables, RNN is the plain vanilla RNN language model (RNN-LM), as defined in The Model section, and LSTM is an LSTM-RNN language model, i.e., an RNN-LM with LSTM memory units.
A-RNN refers to our main model as defined in the Attention in RNN-LM section.
HRED in Table~\ref{tab:movie-test} is the hierarchical neural dialogue model proposed by~\myciteauthoryear{serban-16}.\footnote{We compare to their best-performing model version, that adopts bidirectional gated-unit RNN (GRU).}
LDA-CONV in Table~\ref{tab:ubuntu-ppl-rank} is proposed
by~\myciteauthoryear{luan-16}, which integrates learned
LDA-topic-proportions into an LSTM language model in order to promote topic-concentration in the generations. 
Both tables demonstrate that the attention-RNN-LM (A-RNN) model achieves the best results reported to-date on these datasets in terms all evaluation metrics. It improves the ability of an RNN-LM to model continuous dialogue conversations, while keeping the model architecture simple.
\begin{table}[t]
    \centering
    {\small
    \caption{Results on the {\it MovieTriples} test set. The HRED results are from Serban et al.\ (2016).}\label{tab:movie-test}
    \begin{tabularx}{0.87\linewidth}{l c c c c}
        \toprule
		Model & PPL & PPL@L & WER & WER@L \\
        \midrule
        RNN & $27.09$ & $26.67$ & $64.10\%$ & $64.07\%$\\
        HRED & $26.81$ & $26.31$ & $63.93\%$ & $63.91\%$\\
        A-RNN & $\mathbf{25.52}$ & $\mathbf{23.46}$ & $\mathbf{61.58\%}$ &  $\mathbf{60.15\%}$\\
        \bottomrule
    \end{tabularx}}
\end{table}
\begin{table}[t]
	\centering
    {\small
    \caption{{\it Ubuntu Troubleshoot}  PPL and recall@N, with LSTM and LDA-CONV results from Luan et al.\ (2016).}\label{tab:ubuntu-ppl-rank}
    \begin{tabularx}{0.775\linewidth}{l c c c}
        \toprule
		Model & PPL & recall@$1$ & recall@$2$\\
        \midrule
        RNN & $56.16$ & $11\%$ & $22\%$\\
        LSTM & $54.93$ & $12\%$ & $22\%$\\
        LDA-CONV & $51.13$ & $13\%$ &  $24\%$\\
        A-RNN & $\mathbf{45.38}$ & $\mathbf{17\%}$ & $\mathbf{30\%}$\\
        \bottomrule
    \end{tabularx}}
\end{table}
\begin{table}[h]
    \centering
    {\small
    \caption{RNN-LM vs.\ RNN-Seq2Seq}\label{tab:lang-seq}
    \begin{tabularx}{0.885\linewidth}{l c c}
        \toprule
		PPL & {\it MovieTriples} & {\it Ubuntu Troubleshoot} \\
        \midrule
        RNN-Seq2Seq & $35.10$ & $104.61$\\
        RNN-LM & $\mathbf{27.81}$ &  $\mathbf{56.16}$\\
        \bottomrule
    \end{tabularx}}
\end{table}
\begin{table}[t]
    \centering
    {\small
    \caption{BLEU score on {\it Ubuntu Troubleshoot}}\label{tab:ubuntu-bleu}
    \begin{tabularx}{0.69\linewidth}{l c c}
        \toprule
		Model & dev BLEU & test BLEU \\
        \midrule
        RNN & $0.1846$ & $0.1692$ \\
        A-RNN & $0.2702$ & $0.3713$ \\
        T-A-RNN & $0.2908$ & $0.3128$ \\
        A-RNN-RR & $0.4696$ & $\mathbf{0.4279}$ \\
        T-A-RNN-RR & $\mathbf{0.4895}$ & $0.3971$ \\
        \bottomrule
    \end{tabularx}}
\end{table}
\begin{figure}[!t]
    \centering
   
    \begin{subfigure}[b]{1.0\linewidth}
        {\scriptsize%
          \begin{tabularx}{1.0\linewidth}{p{25px} X}
            \cmidrule{2-2}
            \multirow{2}{*}{History:} & A: person , but this time i
                                         got the money .\\
                                      & R: if you give it to me , i might forget i found you .\\
            \cmidrule{2-2}
            A-RNN: & A:  i ' ll give it to you .\\
            RNN: & A: i don ' t know . \\
            Reference: & A:  i don ' t have it with me . tell person . 
          \end{tabularx}}
        \caption{{\it MovieTriples}\label{output:movie}}     
    \end{subfigure}
    \begin{subfigure}[b]{1.0\linewidth}
        {\scriptsize%
          \begin{tabularx}{1.0\linewidth}{p{25px} X}
            \cmidrule{2-2}
            \multirow{3}{*}{History:} & A: i have really choppy
                                             streaming video in 9.10, any
                                             way to fix that?\\
                                      & A: any way to fix choppy streaming video in 9.10 ?\\
                                      & R: what kind of video\\
            \cmidrule{2-2}
            A-RNN: & A: what video card do you have ?\\
            RNN: & A: what are you trying to do ?\\
            Reference: & A:  what cpu do you have
          \end{tabularx}}
        \caption{{\it Ubuntu Troubleshoot}} 
        \label{output:ubuntu} 
    \end{subfigure}
    \caption{Examples of responses generated by the A-RNN.}
    \label{fig:output}
\end{figure}
\begin{table}[h]
    \centering
    {\small
    \caption{Generation Diversity Results: A-RNN vs.\ RNN}\label{tab:div}
    \begin{tabularx}{0.825\linewidth}{l c c}
        \toprule
		Distinct-1 & {\it MovieTriples} & {\it Ubuntu Troubleshoot} \\
        \midrule
        RNN & $0.0004$ & $0.0007$\\
        A-RNN & $\mathbf{0.0028}$ &  $\mathbf{0.0104}$\\
        \bottomrule
    \end{tabularx}}
\end{table}

We also evaluate the effectiveness of the RNN-LM and RNN-Seq2Seq models on both the {\it MovieTriples} and {\it Ubuntu Troubleshoot} development sets. As shown in Table~\ref{tab:lang-seq}, the RNN language model yields lower perplexity than the RNN sequence-to-sequence model on both datasets. Hence, we present all primary results on our primary A-RNN attention-based RNN language model.\footnote{Experiments also demonstrate significant improvements for the Attention-RNN-LM over the Attention-RNN-Seq2Seq.}

\subsection{Generation Diversity Results}
\label{sec:gen-diversity}
Next, we investigate the ability of the A-RNN to promote diversity in the generations, compared to that of the vanilla RNN using the Distinct-1 metric proposed by~\myciteauthoryear{li-16-diversity}. Distinct-1 is computed as the number of distinct unigrams in the generation scaled by the total number of generated tokens. 
Table~\ref{tab:div} shows that our attention-based RNN language model (A-RNN) yields much more diversity in its generations as compared to the vanilla RNN baseline.

\subsection{Topic Coherence Results}
\label{sec:topic-coherence}
Next, we investigate the ability of the different models to promote topic coherence in the generations in terms of BLEU score. In addition to the RNN and A-RNN models, we consider T-A-RNN, a method that incorporates LDA-based topic information into an A-RNN model, following the approach of \myciteauthoryear{luan-16}. 
We also evaluate our LDA-based re-ranker, A-RNN-RR, which re-ranks according to the score \mbox{$\bar{\mathcal{S}}_m = \lambda \mathcal{S}_m + (1-\lambda) \ell(c_m|w_{0:T})$},
where we compute the log-likelihood $\ell(c_m|w_{0:T})$ based upon a trained A-RNN-M model and validate the weight $\lambda$ on the development set. We also consider a method that combines the T-A-RNN model with an LDA-based re-ranker (T-A-RNN-RR).\footnote{Since~\myciteauthoryear{luan-16} do not publish BLEU scores or implementations of their models, we can not compare with LDA-CONV on BLEU.
Instead, we demonstrate the effect of adding the key component of LDA-CONV on top of the A-RNN.} 
Table~\ref{tab:ubuntu-bleu} reports the resulting BLEU scores for each of these methods on the development and test sets from the {\it Ubuntu Troubleshoot} dataset.
We make the following observations based upon these results:
(1) 
The A-RNN performs substantially better than the RNN with regards to BLEU;
(2) 
using our LDA-based re-ranker further improves the performance by a significant amount (A-RNN v.s.\ A-RNN-RR);
(3)
as opposed to our LDA-based re-ranker, adopting the LDA design of~\myciteauthoryear{luan-16} only yields marginal improvements on the development set, but does not generalize well to the test set (A-RNN v.s.\ T-A-RNN and A-RNN-RR v.s.\ T-A-RNN-RR). Also, our LDA re-ranker results in substantial improvements even on top of their topic-based model (T-A-RNN v.s.\ T-A-RNN-RR).

\subsection{Preliminary Human Evaluation}

In addition to multiple automatic metrics, we also report a preliminary human evaluation.
On each dataset, we manually evaluate the generations of both the A-RNN and RNN models on $100$ examples randomly sampled from the test set.
For each example, we randomly shuffle the two response generations, anonymize the model identity, and ask a human annotator to choose which response generation is more topically coherent based on the conversation history. 
As Table~\ref{tab:human-eval} shows, the A-RNN model wins substantially more often than the RNN model.

\begin{table}[h]
    \centering
    {\small
    \caption{Human Evaluaton: A-RNN vs.\ RNN}\label{tab:human-eval}
    \begin{tabularx}{0.995\linewidth}{l c c}
        \toprule
		 & {\it MovieTriples} & {\it Ubuntu Troubleshoot} \\
        \midrule
        Not distinguishable& $48\%$ & $74\%$\\
        RNN wins & $6\%$ & $5\%$\\
        A-RNN wins & $\mathbf{46\%}$ &  $\mathbf{21\%}$\\
        \bottomrule
    \end{tabularx}}
\end{table}

\subsection{Qualitative Analysis}

Next, we qualitatively evaluate the effectiveness of our A-RNN model through visualizations of the attention and outputs on both datasets.
Figure~\ref{fig:att-viz} provides a visualization of the attention for
a subset of the words in the generation for the two datasets. The last
line in both \figref{att:movie} and \figref{att:ubuntu} presents the
generated response and we highlight in bold two output words (one on
the left and one on the right) for two time steps. For each
highlighted generated word, we visualize the attention weights for
words in the conversation history (i.e., words in the preceding turns
and those previously generated in the output response), where darker
shades indicate larger attention weights. As the figure indicates, the
attention mechanism helps learn a better RNN language model that
promotes topic coherence, by learning to associate the currently-generated word with informative context words in the conversation history.
As shown in Figure~\ref{output:movie}, the A-RNN generates meaningful and topically coherent responses on the {\it MovieTriples} dataset. In comparison, the vanilla RNN tends to produce generic answers, such as ``i don't know''. Similarly, the A-RNN follows up with useful questions on the {\it Ubuntu Troubleshoot} dataset (Fig.~\ref{output:ubuntu}).


\section{Conclusion}
\label{sec:conclusion}

We investigate how to improve the performance of a recurrent neural network dialogue model via an attention mechanism, and how to promote topic and saliency aware conversation continuation. 
Our attention-RNN language model increases the scope of attention continuously as the conversation progresses (which distinguishes it from standard attention with fixed scope in a sequence-to-sequence models) such that each generated word can be associated with its most related words in the conversation history. 
We evaluate this simple model on two large dialogue datasets ({\it MovieTriples} and {\it Ubuntu Troubleshoot}), and achieve the best results reported to-date on multiple dialogue metrics (including complementary diversity-based metrics), performing better than gate-based RNN memory models.
We also promote topic concentration by adopting LDA-based reranking, further improving performance.

\section*{Acknowledgments}

We thank Iulian Serban, Yi Luan, and the anonymous reviewers for sharing their datasets and for their helpful discussion. We thank NVIDIA Corporation for donating GPUs used in this research.


\appendix
\section{Appendix}
\subsection{LDA-based Re-Ranking Details} \label{lda_details}

While the trained attention-RNN dialogue model generates a natural language continuation of a conversation while maintaining topic concentration by token saliency/association, some dialogue-level topic-supervision can help to encourage generations that are more topic-aware.
Such supervision is not commonly available, and we use unsupervised methods to learn document-level latent topics, and adopt the learned topic model to select the best continuation based on document-level topic-matching. 

We choose Latent Dirichlet Allocation (LDA)~\cite{blei-03-lda,blei-09-topic}
due to its demonstrated ability to learn a distribution over latent topics given a collection of documents. 
This generative model assumes that documents arise from multiple topics, each of which is defined as a distribution over a fixed vocabulary of terms. 
Specifically, we can assume that $K$ topics are associated with a collection of dialogues $\{w_{0:T_n}\}_{n=1}^{N}$ in the dialogue corpus, and that each dialogue exhibits these topics with different likelihoods $\theta_{n} \in \mathbb{R}^{K}$. 
Let $\text{Dir}_V(\eta)$ and $\text{Dir}_K(\xi)$ denote a $V$- and $K$-dimensional Dirichlet distributions with scalar parameter $\eta$ 
and vector parameter $\xi \in \mathbb{R}^{K}$, respectively. 
The associated $\eta$ and $\xi$ can be learned by various methods~\cite{blei-03-lda,blei-09-topic,hoffman-10-online}, and the topic proportions
$\hat{\theta}_n$ of a (possibly unseen) dialogue $w_{0:T_n}$ can be estimated with the learned topic structure $\mathcal{L}$. 
We let $\hat{\theta} = \mathcal{L}(w_{0:T})$ denote 
the estimated topic proportions $\hat{\theta}$ of any dialogue $w_{0:T}$ using the learned topic structure $\mathcal{L}$.

We use the training set to learn the topic structure $\mathcal{L}$ with $K$ topics.
Having generated a set of continuations $\{c_m\}_{m=1}^M$ for each unseen dialogue $w_{0:T}$ during testing, we use the topic structure $\mathcal{L}$ to compute the topic proportions for the dialogue and its continuations, $\hat{\theta}(w_{0:T}) = \mathcal{L}(w_{0:T})$ and $\hat{\theta}(c_m) = \mathcal{L}(c_m)$, respectively. Next, we employ a pre-defined metric $\mathcal{S}_m = \mathcal{S}\left(\hat{\theta}(w_{0:T}), \hat{\theta}(c_m)\right)$ to compute the similarity between the topic proportions for the dialogue $\hat{\theta}(w_{0:T})$ and each continuation $\hat{\theta}(c_m)$.
Finally, we compute a weighted score
\mbox{$\bar{\mathcal{S}}_m = \lambda \mathcal{S}_m + (1-\lambda) \ell(c_m|w_{0:T})$}, 
where $\lambda \in [0, 1]$
and $\ell(c_m|w_{0:T})$ is the log-likelihood of the continuation $c_m$ conditioned on the dialogue $w_{0:T}$.
The hyper-parameters $K$ and $\lambda$ are tuned on a held-out development set according to the target evaluation metrics (defined shortly).

Concurrent with our work,~\myciteauthoryear{luan-16} also propose using learned topic representations $\hat{\theta}$ of the given conversation as an extra feature in a language model in order to enhance the topic coherence of the generation.
They define the function $g$ as 
\mbox{$g(h_t, \hat{\theta}, v_j) = O_{v_j}^\top (O_h h_t + O_{\theta} \hat{\theta})$}.
As we show in the Results and Analysis section,
our attention-RNN language models significantly outperform this approach, though it is possible to further improve the quality of the generation by combining these methods i.e., by defining \mbox{$g(h_t, z_t, \hat{\theta}, v_j) = O_{v_j}^\top (O_h h_t + O_z z_t + O_{\theta} \hat{\theta})$} and also using an LDA-based re-ranker. 

\subsection{Dataset Details}

We train and evaluate the models on two large natural language
dialogue datasets, {\it MovieTriples} (pre-processed
by~\myciteauthoryear{serban-16}) and {\it Ubuntu Troubleshoot}
(pre-processed by~\myciteauthoryear{luan-16}). The dialogue within
each of these datasets consists of a sequence of utterances (turns),
each of which is a sequence of tokens
(words). Table~\ref{tab:dataset-stats} provides the statistics for
these two datasets.\footnote{Following the approach of
  \myciteauthoryear{luan-16}, we randomly sample nine utterances as
  negative examples of the last utterance for each conversation in the
  {\it Ubuntu Troubleshoot} for the development set.}
\begin{table}[!h]
    \centering
    {\small
    \caption{Dataset Statistics}\label{tab:dataset-stats}
    \begin{tabularx}{0.8\linewidth}{l r r}
        \toprule
		 & {\it MovieTriples} & {\it Ubuntu} \\
        \midrule
		training size & $196308$ &  $216129$\\
		dev size & $24717$ & $13522$ \\
		test size & $24271$ & $10020$ \\
		turns / dialogue & $3$  & $6$--$20$ \\
		avg. tokens / dialogue & $53$  & $116$ \\
		vocabulary size & $10003$ & $20003$ \\
        \bottomrule
    \end{tabularx}}
 \vspace{-0.25cm}
\end{table}
%


\subsection{Evaluation Metrics} \label{Evaluation Metrics}

For the sake of comparison, we closely follow previous work and adopt several standard (and complementary) evaluation metrics (perplexity, word error rate, recall@N and BLEU, diversity-based Distinct-1).
On the {\it MovieTriples} dataset, we use perplexity (PPL) and word error rate (WER) as is done in previous work, due to the metrics' ability to appropriately quantify the performance of a language model by measuring its ability to learn the syntactic structure of dialogue. Following \myciteauthoryear{serban-16}, we adopt two versions for each metric:
i) PPL as the word-level perplexity over the entire dialogue conversation; 
ii) PPL@L as the word-level perplexity over the last utterance of the conversation; 
iii) WER as the word error rate over the entire dialogue conversation; 
and 
iv) WER@L as the word error rate over the last utterance of the conversation. 
We use all four metrics for model selection on the development set and for
evaluation on the test set.

On the {\it Ubuntu Troubleshoot} dataset, we follow previous work and use PPL and recall@N to evaluate the performance of the proposed language models. 
recall@N~\cite{manning-08-introduction} evaluates a model by
measuring how often the model ranks the correct dialogue continuation within top-N range given $10$ candidates.
Additionally, we also adopt BLEU score~\cite{papineni-01}, to evaluate the quality of generations produced by the models. While not perfect, we believe that BLEU score is a suitable metric for evaluating topic-concentration, as it evaluates the generated continuation by counting explicit n-gram match relative to a set of references, and because word-matching can be a reasonable proximity for topic-matching.

Finally, we also present evaluation using the diversity-based Distinct-1 metric (proposed by~\myciteauthoryear{li-16-diversity}) to measure the ability of the A-RNN to promote diversity in the generations, because typical neural dialogue models generate generic, safe responses (technically appropriate but not informative, e.g., ``I donÕt know''). Distinct-1 is computed as the number of distinct unigrams in the generation scaled by the total number of generated tokens.

\subsection{Training Details} \label{Training Details}

For the {\it MovieTriples} dataset, we follow the same procedure as~\myciteauthoryear{serban-16} and first pretrain on a large {\it Q-A SubTitle} dataset
~\cite{ameixa2014luke} 
and then fine-tune on the target {\it MovieTriples} dataset.
The {\it Q-A SubTitle} dataset contains approximately $5.5$M question-answer pairs, from which we randomly sample $20000$ pairs as the held-out set. 
We performed early-stopping according to the PPL score on the held-out set.
We tuned the dimensionality $d \in \{ 200, 300, 400, 500 \}$ of the RNN hidden states, and chose $d=300$.
We similarly use a setting of $d=300$ on the {\it Ubuntu Troubleshoot} dataset. 
We use Adam~\cite{kingma-15} for optimization in RNN back-propagation.

When training the attention-RNN dialogue model, we perform early-stopping and model selection based on the PPL values on the development set, 
following~\myciteauthoryear{luan-16}. 
For the LDA-based re-ranker, we choose the two hyper-parameters, i.e., the number of topics $K$ and ranking weight $\lambda$, based on the values of recall@N and BLEU score on the development set. 
In the end, we choose $K=10$ out of $K \in \{5, 10, 20, 50\}$ and $\lambda=0.45$ out of $\lambda \in \{0.00, 0.05, \ldots, 0.95, 1.00\}$.

\subsection{Output Examples} \label{Output Examples}
\begin{figure}[!t]
\centering
    \begin{subfigure}[b]{1.0\linewidth}
    \begin{minipage}[t]{1.0\linewidth}
    \begin{minipage}[!c]{0.15\linewidth}
     	{\scriptsize History:}
     \end{minipage}
     \begin{minipage}[!t]{0.85\linewidth}
     	\begin{mdframed}[align=center]{\scriptsize
        	A: i knew how much you wanted it . \\
           R: but i was going to buy it for myself .}
       	\end{mdframed}
      \end{minipage}
      \begin{minipage}[!c]{0.15\linewidth}
           {\scriptsize A-RNN:}
      \end{minipage}
      \begin{minipage}[!t]{0.85\linewidth}
           \begin{mdframed}[align=center,hidealllines=true]
           {\scriptsize
            A: i ' d like to buy you a drink .}
      \end{mdframed}
      \end{minipage}
      \begin{minipage}[!c]{0.15\linewidth}
           {\scriptsize RNN:}
      \end{minipage}
      \begin{minipage}[!t]{0.85\linewidth}
           \begin{mdframed}[align=center,hidealllines=true]
           {\scriptsize
            A: i don ‘ t know . }
      \end{mdframed}
      \end{minipage}
      \begin{minipage}[!c]{0.15\linewidth}
           {\scriptsize Reference:}
      \end{minipage}
      \hfill
      \begin{minipage}[!t]{0.85\linewidth}
           \begin{mdframed}[align=center,hidealllines=true]
           {\scriptsize
            A: i wanted to give it to you . for our anniversary . we met five years ago today .}
      \end{mdframed}
      \end{minipage}
    \begin{minipage}[!c]{0.15\linewidth}
     	{\scriptsize History:}
     \end{minipage}
     \begin{minipage}[!t]{0.85\linewidth}
     	\begin{mdframed}[align=center]{\scriptsize
        	A: person , but this time i got the money .\\
           R: if you give it to me , i might forget i found you .}
       	\end{mdframed}
      \end{minipage}
      \begin{minipage}[!c]{0.15\linewidth}
           {\scriptsize A-RNN:}
      \end{minipage}
      \begin{minipage}[!t]{0.85\linewidth}
           \begin{mdframed}[align=center,hidealllines=true]
           {\scriptsize
            A: i ' ll give it to you .}
      \end{mdframed}
      \end{minipage}
      \begin{minipage}[!c]{0.15\linewidth}
           {\scriptsize RNN:}
      \end{minipage}
      \begin{minipage}[!t]{0.85\linewidth}
           \begin{mdframed}[align=center,hidealllines=true]
           {\scriptsize
            A: i don ' t know .  }
      \end{mdframed}
      \end{minipage}
      \begin{minipage}[!c]{0.15\linewidth}
           {\scriptsize Reference:}
      \end{minipage}
      \begin{minipage}[!t]{0.85\linewidth}
           \begin{mdframed}[align=center,hidealllines=true]
           {\scriptsize
            A: i don ' t have it with me . tell person .}
      \end{mdframed}
      \end{minipage}
    \end{minipage}
    \caption{{\it MovieTriples}\label{output:movie}}     
    \vspace{5pt}
    \end{subfigure}
    \begin{subfigure}[b]{1.0\linewidth}
    \begin{minipage}[t]{1.0\linewidth}
    	\begin{minipage}[!c]{0.15\linewidth}
     	{\scriptsize History:}
     \end{minipage}
     \begin{minipage}[!t]{0.85\linewidth}
     	\begin{mdframed}[align=center]{\scriptsize
        	A: so i need to make an ntfs partition .... help ?\\
           R: install ntfsprogs \\
           R: it will add the functionality to gparted}
       	\end{mdframed}
      \end{minipage}
      \begin{minipage}[!c]{0.15\linewidth}
           {\scriptsize A-RNN:}
      \end{minipage}
      \begin{minipage}[!t]{0.85\linewidth}
           \begin{mdframed}[align=center,hidealllines=true]
           {\scriptsize
            R: sudo apt-get install ntfsprogs}
      \end{mdframed}
      \end{minipage}
      \begin{minipage}[!c]{0.15\linewidth}
           {\scriptsize RNN:}
      \end{minipage}
      \begin{minipage}[!t]{0.85\linewidth}
           \begin{mdframed}[align=center,hidealllines=true]
           {\scriptsize
            R: how do i do that ?}
      \end{mdframed}
      \end{minipage}
      \begin{minipage}[!c]{0.15\linewidth}
           {\scriptsize Reference:}
      \end{minipage}
      \begin{minipage}[!t]{0.85\linewidth}
           \begin{mdframed}[align=center,hidealllines=true]
           {\scriptsize
            R: sudo apt-get update ; sudo apt-get install ntfsprogs}
      \end{mdframed}
      \end{minipage}
     \begin{minipage}[!c]{0.15\linewidth}
     	{\scriptsize History:}
     \end{minipage}
     \begin{minipage}[!t]{0.85\linewidth}
     	\begin{mdframed}[align=center]{\scriptsize
        	A: i have really choppy streaming video in 9.10 , any way to fix that ?\\
        	A: any way to fix choppy streaming video in 9.10 ?\\
           R: what kind of video}
       	\end{mdframed}
      \end{minipage}
      \begin{minipage}[!c]{0.15\linewidth}
           {\scriptsize A-RNN:}
      \end{minipage}
      \begin{minipage}[!t]{0.85\linewidth}
           \begin{mdframed}[align=center,hidealllines=true]
           {\scriptsize
            R: what video card do you have ?}
      \end{mdframed}
      \end{minipage}
      \begin{minipage}[!c]{0.15\linewidth}
           {\scriptsize RNN:}
      \end{minipage}
      \begin{minipage}[!t]{0.85\linewidth}
           \begin{mdframed}[align=center,hidealllines=true]
           {\scriptsize
            R: what are you trying to do ?}
      \end{mdframed}
      \end{minipage}
      \begin{minipage}[!c]{0.15\linewidth}
           {\scriptsize Reference:}
      \end{minipage}
      \begin{minipage}[!t]{0.85\linewidth}
           \begin{mdframed}[align=center,hidealllines=true]
           {\scriptsize
            R: what cpu do you have}
      \end{mdframed}
      \end{minipage}
    \end{minipage}
    \caption{{\it Ubuntu Troubleshoot}} 
    \label{output:ubuntu} 
    \vspace{5pt}
    \end{subfigure}
    \caption{Examples of responses generated by the A-RNN along with
      the reference response.}
    \vspace{-7pt}
    \label{fig:output}
\end{figure}
We qualitatively evaluate the effectiveness of our A-RNN model through visualizations of the output on both datasets.
The A-RNN generates meaningful and topically coherent responses on the
{\it MovieTriples} dataset (Fig.~\ref{output:movie}). In comparison, the vanilla RNN tends to produce generic answers, such as ``i don't know''.  On the {\it Ubuntu Troubleshoot} dataset, the A-RNN either provides promising technical solutions or follows up with useful questions (Fig.~\ref{output:ubuntu}).

{\footnotesize
\bibliographystyle{aaai}
\bibliography{references}
}
\end{document}